\documentclass[lettersize,journal]{IEEEtran}
\usepackage{amsmath,amsfonts}
\usepackage{algorithmic}
\usepackage{algorithm}
\usepackage{array}
\usepackage[caption=false,font=normalsize,labelfont=sf,textfont=sf]{subfig}
\usepackage{textcomp}
\usepackage{stfloats}
\usepackage{url}
\usepackage{amsmath, amsfonts}
\usepackage{verbatim}
\usepackage{graphicx}
\usepackage{bm}
\usepackage{authblk}
\usepackage{biblatex}
\newcommand{\subsubsubsection}[1]{\paragraph{#1}\mbox{}\\}
\begin{document}

\title{6-DOF All-Terrain Cyclocopter}

\author[a]{Jingwei Li}
\author[b]{Boyuan Deng}
\author[a]{Xinyu Zhang \thanks{Corresponding author: xyzhang@tsinghua.edu.cn}}
\author[a]{Kangyao Huang}
\affil[b]{National Key Lab of Autonomous Intelligent Unmanned Systems, School of Automation, Beijing Institute of Technology University}
\affil[a]{Mengshi Intelligent Vehicle Team, School of Vehicle and Mobility, Tsinghua University}


\renewcommand*{\Affilfont}{\small\it} 
\renewcommand\Authands{ and } 
\date{} 


\maketitle
\begin{abstract}
This paper presents the design of a 6-DOF all-terrain micro aerial vehicle and two control strategies for multimodal flight, which are experimentally validated. The micro aerial vehicle is propelled by four motors and controlled by a single servo for the control of the cycloidal rotors(cyclorotors) speed and lift direction. Despite the addition of the servo, the system remains underactuated. To address the traditional underactuation problem of cycloidal rotor aircraft, we increase the number of control variables. We propose a PID and a nonlinear model predictive control (NMPC) framework to tackle the model's nonlinearities and achieve control of attitude, position, and their derivatives.Experimental results demonstrate the effectiveness of the proposed multimodal control strategy for 6-DOF all-terrain micro aerial vehicles. The vehicle can operate in aerial, terrestrial, and aquatic modes and can adapt to different terrains and environmental conditions. Our approach enhances the vehicle's performance in each mode of operation, and the results show the advantages of the proposed strategy compared to other control strategies. 
\end{abstract}


\section{Introduction}
\IEEEPARstart{M}{ultimodal} micro air vehicles (MMAVs) are a new class of drones that have recently emerged, thanks to advancements in microelectronics, materials sciencer\cite{TiltDrone}, and aviation\cite{qin2018vins}, \cite{zhou2021raptor}. Inspired by birds, insects, and other creatures in nature, multimodal aerial vehicles are capable of adapting to diverse environments during mission execution. By switching between different modes of motion, these vehicles can minimize energy consumption and cope with more complex environments, while exhibiting high agility, efficiency, flexibility, and safety. 

One of the most notable advantages of multimodal aerial vehicles is their versatility. They can easily traverse challenging environments, making them ideal for tasks such as surveillance, reconnaissance, search and rescue, infrastructure inspection, and precision parcel delivery. Additionally, portability and durability are key advantages of multimodal aerial vehicles. Due to their lightweight structure and high-capacity lithium-polymer batteries, they can be quickly transported to remote areas and fly for extended periods. This makes them well-suited for emergency response scenarios and environmental monitoring applications.

\begin{figure*}[!t]
\centering
\subfloat[HyTAQ Vehicle]{\includegraphics[width=2in]{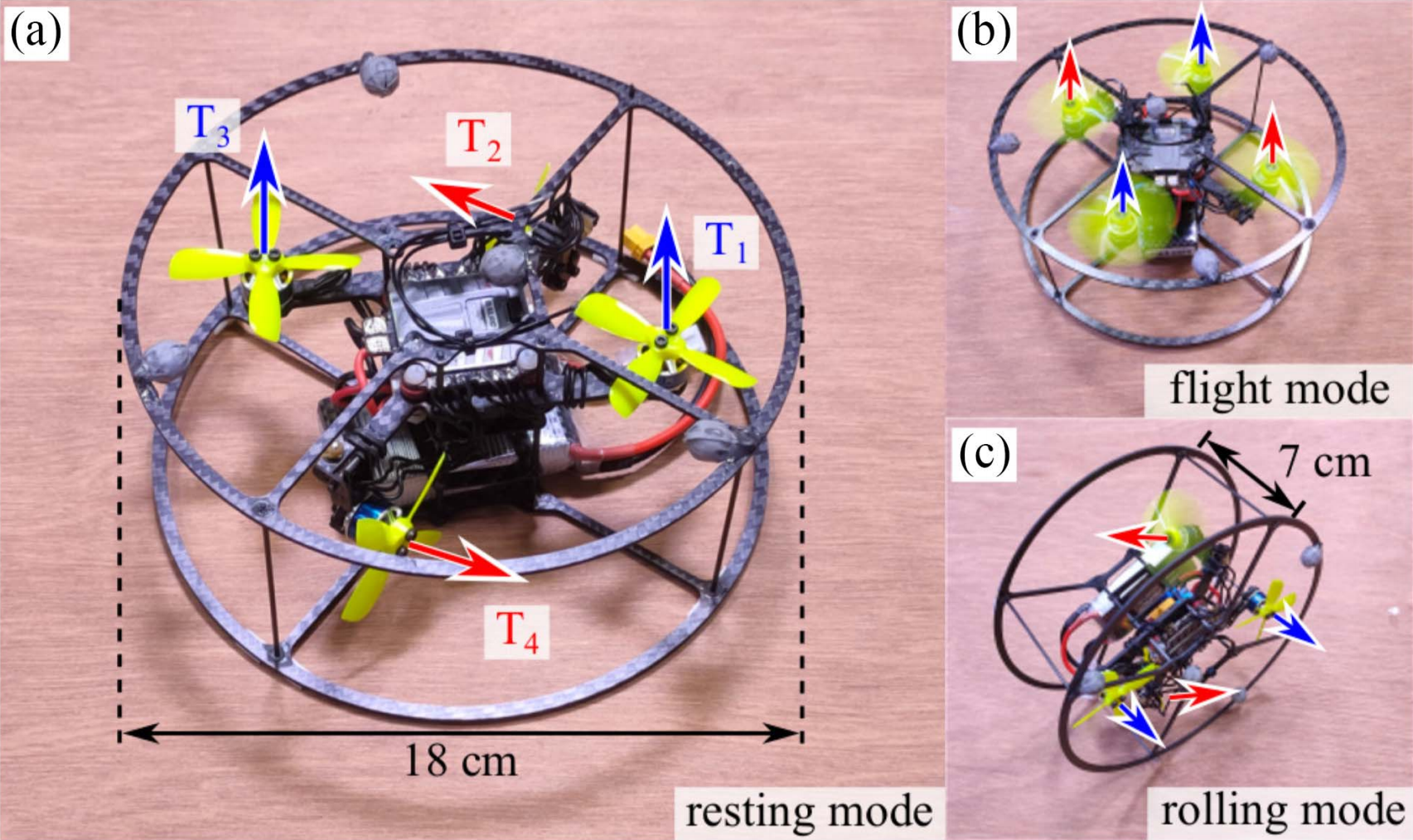}%
\label{Fig:HyTA}}
\hfil
\subfloat[Bipedal Robot]{\includegraphics[width=2in]{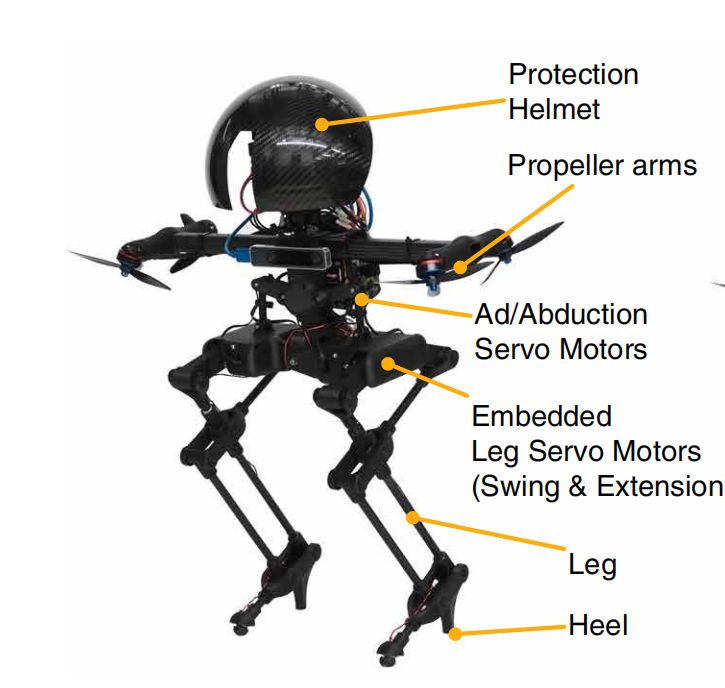}%
\label{Fig:BiLeg}
}
\hfil
\subfloat[Nezha]{\includegraphics[width=2in]{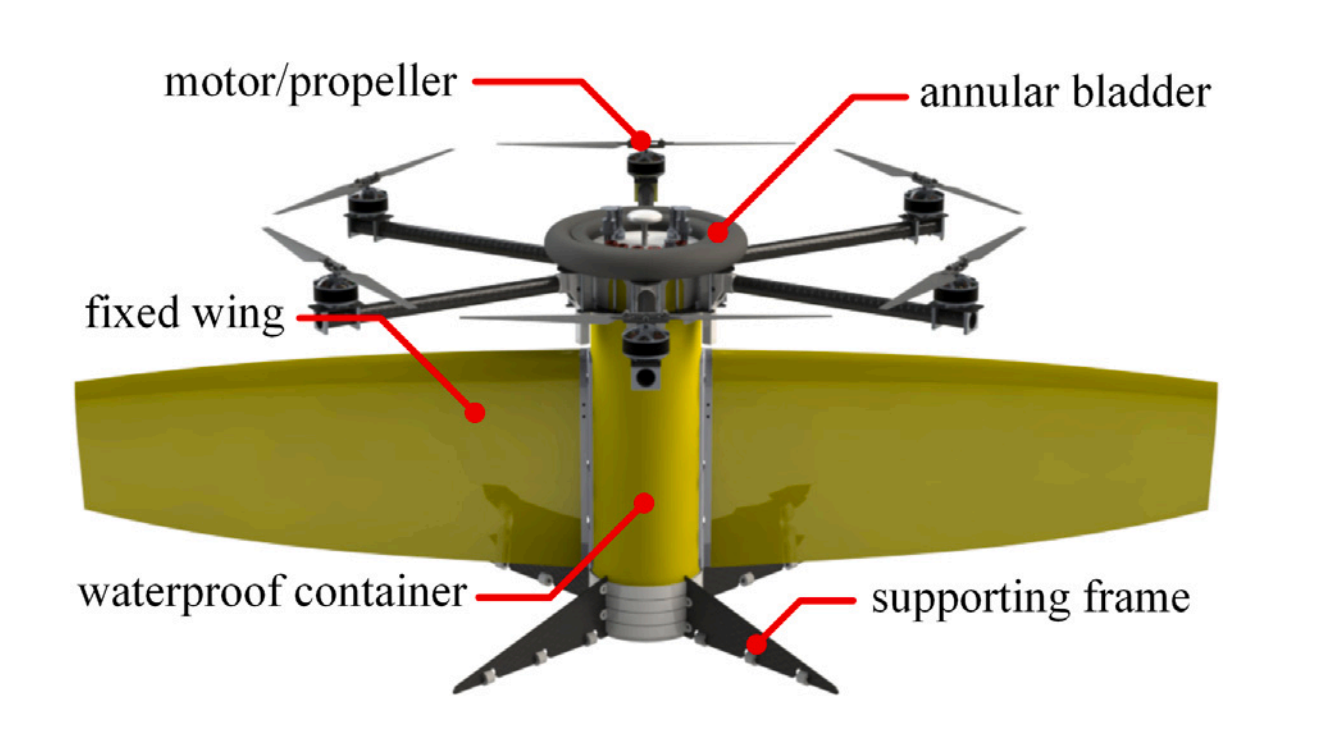}%
\label{Fig:aquN}
}
\caption{Existing multimodal air vehicles.}
\end{figure*}

The multimodal mobility capability of aerial vehicles has a wide range of potential applications, such as maritime patrol, search and rescue, mountain search and rescue, disaster relief, and more. Future research and application of multimodal aerial vehicles are expected to further promote their development.Rotary-winged multimodal MAV such as the hybrid terrestrial and aerial quadrotor (HyTAQ) by utilizing traditional edgewise rotors to generate thrust and installing external wheels on a conventional quadrotor, the platform (Fig.\ref{Fig:HyTA}) with a total weight of 156g can passively deform its body by controlling the rotation speed, enabling the transition between ground and aerial modes. Although the mechanical design in the multimodal configuration is straightforward, the ground mode still relies on lift generated by rotation instead of direct torque produced by the external wheels, which increases the complexity of the control strategy and reduces the agility of the system.

MMAVs inspired by biological systems, such as the morphing micro air-land vehicle (MMALV)\cite{boria2005sensor}, \cite{kim2021bipedal}, \cite{boria2005sensor}achieve flight using a combination of nose propeller and morphing flexible wings, despite being able to fly and move on the ground, the agility of the MAV is limited. In contrast, \cite{kim2021bipedal} achieves the transition between ground and aerial modes by equipping a bipedal robots with distributed electric thrusters, which has a weight of 2.58 kg and an overall height of 75 cm (Fig. \ref{Fig:BiLeg}), demonstrates excellent mobility in ground mode, but the weight of it limits the duration of ground operations, and reduces the agility during flight missions.

Multi-modal UAVs also include hybrid aerial underwater vehicle (HAUVs), such as Nezha\cite{lu2021water}, which by combining the structure of fixed-wing and rotary-wing UAVs and utilizing an aerodynamic buoyancy system, Nezha, weighing 18 kg, is capable of achieving multiple modes such as hovering in the air and gliding underwater while maintaining a balance between flight payload and underwater weight. However, due to the rotary-wing propulsion system, Nezha lacks a ground mode, and the presence of motor arms may cause instability in underwater performance.

In the design of micro air vehicles, fluid viscous forces dominate due to their small size, and inertial forces are relatively weak. Therefore, the Reynolds number of micro air vehicles is relatively low, and the current challenge is to design models with high lift-to-drag ratios at low Reynolds numbers. As rotary-wing aircraft require constant changes in rotor speed and blade angle to maintain stable flight, this leads to significant mechanical energy losses and aerodynamic drag, further reducing its lift-to-drag ratio. Although \cite{chin2020efficient} has solved the problems of flight efficiency and endurance, it cannot perform special maneuvers such as hovering. Another type of wing configuration is the cycloidal-rotor(cyclorotor). The blades on a conical cyclorotor rotate around a horizontal axis, causing the blades to span over the rotation axis parallel to the flight direction and perpendicular to it. In order to generate net lift, the angle of each blade is cyclically changed during rotation via a four-bar linkage mechanism, causing each blade's top and bottom sections to produce positive geometric pitch angle simultaneously during hover \cite{benedict2013effect}. The magnitude of the lift vector provided is controlled by the rotation speed.

The use of a reciprocating rotor as a rotor system has many advantages in terms of its unique blade design and four-bar linkage structure, including aerodynamic efficiency, multi-modal maneuverability, and climb rate. Previous research\cite{benefit1} has shown that reciprocating rotors have higher power loading(thrust/power) capabilities.

Due to the ability of the rotor to rotate the lift vector instantly through phase cyclic pitch, the Cyclogyro aircraft can maintain a horizontal pitch angle during forward flight, thereby reducing parasitic drag and power consumption during forward flight\cite{jarugumilli2014wind}. Subsequent studies have explored many configurations based on cyclorotor design \cite{ yun2005thrust,yun2007design,hwang2008development}. A cyclorotor aircraft with two cycloidal rotors as main thrusters and a conventional nose rotor has been shown to achieve a maximum horizontal flight speed of 7m/s\cite{benefit2}. Moreover, the multi-modal capability of the cyclorotor aircraft has been demonstrated by adding external wheels for land-air-water mode transitions\cite{shrestha2021all}. However, due to the nature of cyclorotor rotation direction, which changes the direction of lift vector by altering rotor speed, instantaneous transitions between flight and hovering modes are not possible.

The motion of a cycloidal rotor at low speeds can be approximated as that of a wheel, which eliminates the need for any additional design changes compared to the Nezha and MMALV platforms to achieve rolling motion on the ground. Therefore, ground motion can be achieved directly through motor torque. Recent work has utilized the advantages of cycloidal rotors to develop a multi-modal cycloidal flying vehicle, demonstrating stable hovering and ground movement\cite{shrestha2021all}. To develop a more versatile and functional multi-modal cycloidal flying vehicle, our work builds upon previous designs and continues to explore further advancements.

\section{Mechanical Design and Mathematical Modelling}

\subsection{Mechanical design}

\subsubsection{Blade kinematics}

\subsubsection{Cyclorotor design}

\subsection{Mathematical modelling}

Based on the actual mechanical model, a 4 - input - 6 - output (MIMO) mathematical model can be constructed. Because the vehicle can move in a variety of media, the dynamic models of aerial, terrestrial and aquatic modes are different. On this basis, the multi-freedom and multi-mode control of the vehicle can be realized.

\subsubsection{Kinematic model}

As for the kinematics model of the vehicle, the main research is the mapping of its own velocity and attitude in the world coordinate system. Therefore, two coordinate systems are defined, namely the body coordinate system (OB) and the Earth coordinate system (OE). The origin of the body coordinate system is defined at the center of mass of the body and will move with the movement of the body. The earth coordinate system is consistent with the Cartesian right-hand coordinate system.By coordinate transformation:
\begin{equation}
    \begin{aligned}
        V_{OE}=T_{oe}^{ob}\cdot V_{OB}\\
T_{oe}^{ob}=\begin{bmatrix}
C_{\psi}C_{\theta} & C_{\psi}S_{\theta}S_{\phi}-S_{\psi}C_{\phi} & C_{\psi}S_{\theta}C_{\phi}-S_{\psi}S_{\phi}\\ 
S_{\psi}C_{\theta} & S_{\psi}S_{\theta}S_{\phi}+C_{\psi}C_{\phi} & S_{\psi}S_{\theta}C_{\phi}-C_{\psi}S_{\phi}\\ 
-S_{\theta} & C_{\theta}S_{\phi} & C_{\theta}C_{\phi}
\end{bmatrix}
    \end{aligned}\label{coordinate transformation  v}
\end{equation}
where \(V_{OE}={[\dot{x} \ \dot{y} \ \dot{z}]}^{T}\) represents linear velocity in the earth coordinates,\(V_{OB}={[u \ v \ w]}^{T}\) represents linear velocity in the body coordinates.\(T_{oe}^{ob}\) represents the transformation matrix from the body coordinate system to the earth coordinate system.In this matrix, \(C\) represents \(cos\), \(S\) represents \(sin\), and \(\phi \ \theta \ \psi \) respectively represent roll,pitch and yaw in Euler angle.

When the vehicle changes its attitude, the corresponding angular velocity can be defined as \(w_{OE} = {[\dot{\phi} \ \dot{\theta} \ \dot{\psi}]}^{T}\). Instead, the corresponding angular velocity change in the body coordinate system can be defined as \(w_{OB} = {[w_{x} \ w_{y} \ w_{z}]}^{T}\).The equation is as follows:
\begin{equation}
    \begin{aligned}
        \begin{bmatrix}
w_{x}\\ w_{y}
\\ w_{z}
\end{bmatrix}=\begin{bmatrix}
1 & 0 &-S_{\theta} \\ 
0 & C_{\phi} &S_{\phi}C_{\theta} \\ 
0 & -S_{\phi} &C_{\phi}C_{\theta} 
\end{bmatrix}\begin{bmatrix}
\dot{\phi}\\ 
\dot{\theta}\\ 
\dot{\psi}
\end{bmatrix}
    \end{aligned}\label{coordinate transformation  w}
\end{equation}
In summary of the above steps, equations 1 and 2 respectively calculate the linear velocity matrix representation and angular velocity expression of the center of mass in the world coordinate system. They together form the kinematics equation expression of cyclorotor, providing the basis for PID and NMPC control.
\subsubsection{Dynamic Model}

\subsubsubsection{Aerial mode}
By increasing the number of control variables, the vehicle is able to achieve 6-DOF control, which includes translational motion along three axes and rotational motion around these axes. We model the vehicle as a rigid body controlled by four cycloidal rotors.The dynamic equations of motion are derived using the Newton-Euler equations, which describe the relationship between forces, torques, and the resulting motion of the vehicle:
\begin{equation}
	\begin{aligned}
		\bm{\dot{p}}_{WB}&=\bm{v}_{WB}                &\bm{\dot{q}}_{WB}&=\frac{1}{2}\bm{\Lambda} (\bm{\omega}_{B}) \cdot \bm{q}_{WB}\\
		\bm{\dot{v}}_{WB}&=\bm{\dot{q}}_{WB}\odot\bm{c}-\bm{g} &\bm{\dot{\omega}}&=\bm{J^{-1}}(\bm{\eta} -\bm{\omega}_B\times \bm{J}\bm{\omega}_B)\\
	\end{aligned}\label{eq:Aerial dynatic}
\end{equation}
where $\bm{p}_{WB}=[p_x, p_y, p_z]^T$ and $\bm{v}_{WB}=[v_x, v_y, v_z]^T$ represent the position and velocity vectors of the vehicle in thr world frame $W$, respectively. We use a unit quaternion $\bm{q}_{WB}=[q_w, q_x, q_y, q_z]^T$ to describe the orientation of the vehicle and use $\bm{\omega}_B=[\omega_x, \omega_y, \omega_z]^T$ to denote the body rates in the body frame $B$. $\bm{g}=[0, 0, g_z]$ with $g_z=9. 81m/s^2$ is the gravity vector, $\bm{J}$ is diagonal inertia matrix, $\bm{\eta}$ is the torque, and $\bm{\Lambda} (\bm{\omega}_{B})$ is a centroskew symmetric matrix. Finally, $\bm{c}=[0, 0, c]^T$ is the mass thrust vector. 
\subsubsubsection{Terrestrial mode}
Switching between aerial and terrestrial modes presents two challenges: handling of landing gear and changing the direction of rotation of cycloidal rotors. The pivot points of the landing gear are designed to be controlled by servos (as shown in Fig. 1), enabling the gear position to be changed programmatically to accommodate the four-wheel differential drive system of the terrestrial mode. In aerial mode, the direction of rotation of the cycloidal rotors is opposite to that of the terrestrial mode. Therefore, to achieve turning and other maneuvers in terrestrial mode, the motors must also be capable of reversing their direction. By selecting programmable electronic speed controllers, the motor direction can be reversed, thus enabling the four-wheel differential drive system. The dynamics of the system are described by the following equations:
\begin{equation}
    \begin{aligned}
        \dot{\mathrm{x}}&=\mathrm{v} \cos \theta \\
        \dot{\mathrm{y}}&=\mathrm{v} \sin \theta \\
        \dot{\theta}&=\frac{\mathrm{v}_{\mathrm{R}}-\mathrm{v}_{\mathrm{L}}}{\mathrm{W}}\\
        \mathrm{v}&=\frac{\mathrm{v}_{\mathrm{R}}+\mathrm{v}_{\mathrm{L}}}{2}
    \end{aligned}\label{eq:Terrestrial dynatic}
\end{equation}
Where $\mathrm{v}_{\mathrm{R}}$ and $\mathrm{v}_{\mathrm{L}}$ represent the linear velocities of the left and right wheels, respectively, and it is assumed that the rotational speeds of the wheels on the same side are equal, $\mathrm{W}$ is the distance between the left and right wheels.
\subsubsubsection{Aquatic mode}
In amphibious mode, conventional cycloidal helicopters cannot transition from flight mode to aquatic mode in a continuous process because their blade design is fixed and the direction of thrust cannot be changed. During flight, the thrust always points downward along the Z-axis. However, our cyclocopter design allows for the direction of thrust to be changed by manipulating the servo. This enables the thrust to change from the Z-axis to the X-axis. To conserve energy, the aquatic mode only uses rear-wheel drive. Therefore, the cyclocopter in aquatic mode can be simplified as a two-wheel differential drive model with the following dynamics equation:
\begin{equation}
    \begin{aligned}
        \dot{\mathrm{x}}&=\mathrm{v} \cos \theta \\
        \dot{\mathrm{y}}&=\mathrm{v} \sin \theta \\
        \dot{\theta}&=\frac{\mathrm{v}}{\mathrm{W}}\tan\phi\\
    \end{aligned}\label{eq:Aquatic dynatic}
\end{equation}
Where $\mathrm{v}$ represents the linear velocity of the vehicle, $\phi$ represents the steering angle of the vehicle, and $\mathrm{W}$ represents the wheelbase of the vehicle, which is the distance between the left and right rotors.

\section{Multimodal Control Strategy}
\subsection{Transition strategy}
In order to demonstrate the controllability of this vehicle in the case of mechanical structure and multi-mode, it is planned to design experiments involving aerial, terrestrial and aquatic modes.However, considering the different dynamic models of the vehicle in the aerial, terrestrial and aquatic mode, different algorithms will be used to calculate the input rotational speed.In order to solve the difference of multi-mode algorithm, finite state machine is used to switch and select three states.

The finite state machine contains three states,corresponding to aerial, terrestrial and aquatic state.The three states all contain different motion states, which are mainly defined according to the set expected path and state switching time. The followings are the various motion states.
\subsubsection{Aerial state}
\subsubsubsection{Static}
Since the driving force of the vehicle comes from the quadrotor, the landing gear needs to be opened when the vehicle is at 
the aerial state, so as to generate upward lift force. This state is mainly connected before takeoff and when landing on the ground.
\subsubsubsection{Takeoff}
In the take-off condition, considering the actual flight situation, not only includes direct flight, usually accompanied by changes in roll,pitch and yaw. Due to the existence of steering gear in the mechanical structure, yaw will not appear uncontrollable situation, and the control of yaw is different from roll and pitch. Due to the particularity of yaw, an angle should be given to the steering gear to balance the torque of yaw.
\subsubsubsection{Hovering}
According to the expected path and the actual path obtained by feedback, judge whether the hover point has been reached. After entering the Hovering, wait for instructions for the next step.
\subsubsubsection{Landing}
Similar to Takeoff, Landing also will change its posture depending on the actual situation. And the vehicle will choose the next step to Hovering or Static, depending on the landing point.
\subsubsection{Terrestrial state}
\subsubsubsection{Static}
Since the driving force of the terrestrial state is in the same direction as the forward direction, the landing gear needs to be removed. So Static of terrestrial state is the landing gear retractable state for the next operation.
\subsubsubsection{Driving}
It can realize the basic operation in the driving process and complete the straight or turn operation according to the target trajectory.
\subsubsection{Aquatic state}
\subsubsubsection{Static}
Since the driving force in the aquatic state is generated in a similar way to the aerial state, it is also necessary to open the landing. Moreover, in order to change the driving force direction to be consistent with the forward direction, it is necessary to set the steering gear angle to be consistent with the direction of movement.
\subsubsubsection{Driving}
Unlike the aerial and aquatic state, the driving in the aquatic state is improved from a four-rotor to a two-rotor to save energy. Therefore, in actual operation, we will not give power to the rotor which is without the steering gear.
\subsection{Control strategy}
In order to verify the controllability of the mechanical structure and the feasibility of multimodal switching, it is necessary to select a controller to control the mechanical structure.Considering the innovation of mechanical structure, PID control is chosen as the baseline. However, due to the nonlinear structure of vehicle itself and the serious coupling phenomenon between controlled objects, nonlinear model predictive control (NMPC) controller is selected as the nonlinear control method of aerial mode based on PID.

\subsubsection{Cascade PID controller}
Referring to the PID control system of traditional vehicle, altitude controller and attitude controller are adopted to solve the nonlinear and coupling problems of the model.Altitude and attitude control can be obtained from the following formula:
\begin{equation}
    \begin{aligned}
\begin{bmatrix}\delta_{x }
\\ \delta_{y }
\\ \delta_{z }
\end{bmatrix}=K_{p}\begin{bmatrix}e_{x}
\\ e_{y}
\\ e_{z}

\end{bmatrix}+K_{i}\int\begin{bmatrix}e_{x}
\\ e_{y}
\\ e_{z}
\end{bmatrix}dt+K_{d}\begin{bmatrix}\dot{e_{x}}
\\ \dot{e_{y}}
\\ \dot{e_{z}}
\end{bmatrix}
    \end{aligned}\label{eq:PID pos}
\end{equation}
\begin{equation}
    \begin{aligned}
\begin{bmatrix}\delta_{\phi }
\\ \delta_{\theta }
\\ \delta_{\psi }
\end{bmatrix}=K_{p}\begin{bmatrix}e_{\phi}
\\ e_{\theta}
\\ e_{\psi}

\end{bmatrix}+K_{i}\int\begin{bmatrix}e_{\phi}
\\ e_{\theta}
\\ e_{\psi}
\end{bmatrix}dt+K_{d}\begin{bmatrix}\dot{e_{\phi}}
\\ \dot{e_{\theta}}
\\ \dot{e_{\psi}}
\end{bmatrix}
    \end{aligned}\label{eq:PID theta}
\end{equation}
where \( k_{p},k_{i},k_{d}\) are proportional gain, integral gain and differential gain.Moreover,\( e_{x}=x_{ref}-x\,e_{y}=y_{ref}-y\,e_{z}=z_{ref}-z\) is error in the x,y,z direction,\( e_{\phi}=\phi_{ref}-\phi,e_{\theta}=\theta_{ref}-\theta,e_{\psi}=\psi_{ref}-\psi\) are 
the errors of roll,pitch and yaw degrees of freedom. 

The system obtains the position and attitude of the current moment, and passes it into the corresponding altitude ring and attitude ring. Then the rotation speed under the corresponding mode is calculated to form a closed-loop PID system.
\begin{figure*}[htbp]
\centering
{\includegraphics[width=1\textwidth]{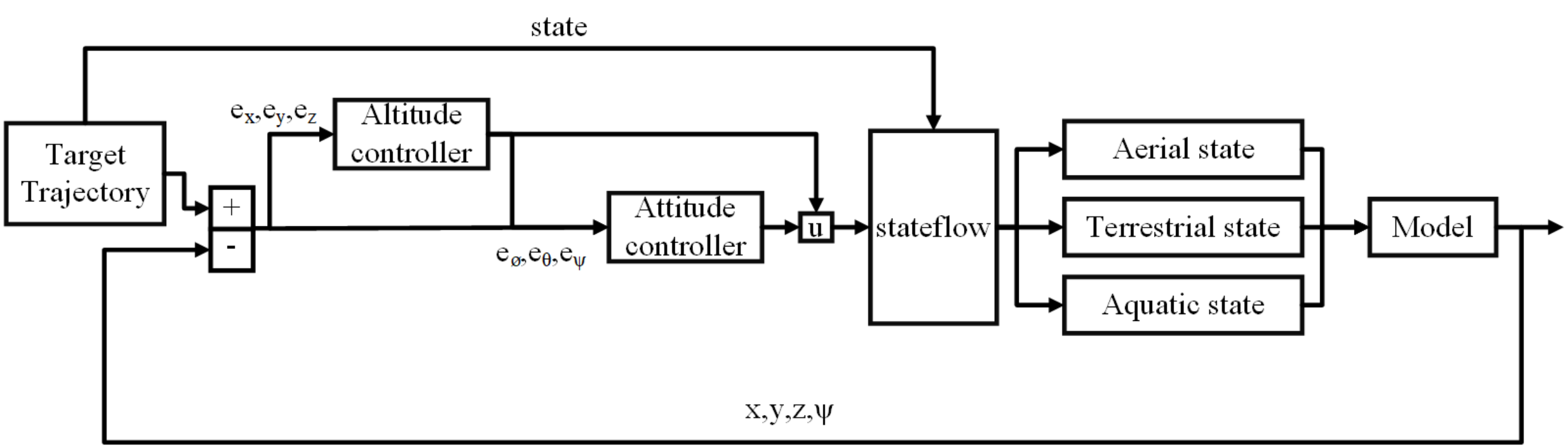}}
\caption{PID closed-loop feedback system}
\label{Fig:PID}
\end{figure*}
\subsubsection{Nonlinear Model Predictive Control (NMPC) controller}
In daily life, when a command is given to an aircraft, the command is often not immediately responded to, but after a delay. This delay is a challenge for different controllers. Due to the defects of PID controller itself, there will be overshoot and more serious shock in the process of response. However, compared with MPC controller, it can handle these problems better. Therefore, in consideration of the accurate handling of response and the nonlinear problems of the model, this paper chooses to optimize PID in the aerial state and uses NMPC controller instead.

The basic idea of model predictive control is to use the known model, the current state of the system, and the future control amount to predict the future output of the system. The control is realized by solving the constrained optimization problem in a rolling way. It has three characteristics: prediction model, rolling optimization and feedback correction. The main body of model prediction controller is mainly composed of linear error model, cost function and system constraint. The error model in this paper is the dynamic model mentioned before, which is the basis of the agent control algorithm. The design of cost function takes into account the rapidity and stationarity of trajectory tracking. System constraints include vehicle actuator constraints, control smoothing constraints and vehicle stability constraints. The theoretical basis of mpc control is to solve the objective function with system constraints according to the error model.

The first is the error model. Since the research object of this paper is the nonlinear system, it needs to adopt the nonlinear error model, which is mentioned in the aerial dynamic model.

In order to account for dynamic feasibility, the cost function needs to be set according to the outputs and the controlled objects.  Therefore, the cost function can be written as:
\begin{equation}
    \begin{aligned}
        J(x_{k},u_{k})=\sum_{j=0}^{N-1}{\underset{tracking\ cost}{\underbrace{{[||y(k+j|k)-y_{ref}(k+j)||_{Q}^{2}}}}+\underset{control\ effort}{\underbrace{||u(k+j|k)||_{R}^{2}}}}
    \end{aligned}\label{cost function}
\end{equation}

where \( x,u,y\) are state variables,input matrix,output matrix. According to the actual situation, the output matrix contains coordinates in the x,y,z directions and yaw, and the input matrix contains thrust and torques in the x,y,z directions.

As for the system constraints, you need to set them according to the actual application scenario.During takeoff and landing, it is necessary to prevent the calculation of the speed of the case of too large or too small. According to the actual use of the motor, the thrust limits are as follows:
\begin{equation}
    \begin{aligned}
        -5m/s^{2}\leq acceleration\leq 15m/s^{2}
    \end{aligned}\label{constraints on thrust}
\end{equation}
For Euler Angle, if the tilt of the fuselage is too large, it will also cause instability of the system. Therefore, in order to prevent the instability of the system during rolling and pitching operations, the following constraints are implemented.
\begin{equation}
    \begin{aligned}
    \begin{split}
        -\pi /6\leq roll\leq \pi /6\\
        -\pi /6\leq pitch\leq \pi /6
        \end{split}
    \end{aligned}\label{constraints on thrust}
\end{equation}
To sum up, the specific problem of NMPC can be achieved by solving the following optimal problem.
\begin{equation}
    \begin{aligned}
        min J(x_{k},u_{k})\\
s.t.\ constraints (7),(8)
    \end{aligned}\label{NMPC}
\end{equation}
Formulation (9) presents a nonlinear programming problem with complementarity constraints, the computational tractability of which has been revealed in previous studies\cite{HyTAQs_via_NMPC}.
\section{Experiment Results}
In order to prove the controllability of the mechanical structure and the feasibility of the control method, we built an ideal mathematical model in Simulink for simulation.
\subsection{Trajectory tracking in mulitimodal operation}
To clearly demonstrate the multiple states of the vehicle in the aerial, terrestrial and aquatic modes, we set the following path based on the x-coordinate points.

Terrestrial medium:\ \( 0m\leq x\leq 100m\)

Aerial medium:\ \( 100m\leq x\leq 200m\)

Aquatic medium:\ \( 200m\leq x\leq 300m\)

The vehicle goes straight from (0,0,0) to (100,0,0). Using this as the take-off point, open the landing gear and prepare for take-off. It flies over (100,0,100), (200,100,150) and hovers on arrival. During descent, it passes (150,80,100) and hovers, landing at (200,0,0). From this point it enters the aquatic medium and eventually travels to (300,100,0).

\subsection{Comparison of NMPC and PID effect}

As can be seen from the figure, PID as baseline control of the cyclocopter can complete the goal of trajectory planning. Based on this, the controllability of the structure is confirmed. Furthermore, we replace the aerial mode controller with NMPC and compare the effect with PID.

The set trajectory of the aerial module remains unchanged, and the coordinates of NMPC and PID on x,y and z change as follows.
 \section{Conclusion}
 

\printbibliography
\end{document}